# Minimizing The Misclassification Error Rate Using a Surrogate Convex Loss

**Shai Ben-David**                                                                                      shai@cs.uwaterloo.ca
University of Waterloo

**David Loker**                                                                                         dloker@cs.uwaterloo.ca
University of Waterloo

**Nathan Srebro**                                                                                       nati@ttic.edu
Toyota Technological Institute at Chicago

**Karthik Sridharan**                                                                                   skarthik@wharton.upenn.edu
University of Pennsylvania

## Abstract

We carefully study how well minimizing convex surrogate loss functions corresponds to minimizing the misclassification error rate for the problem of binary classification with linear predictors. We consider the agnostic setting, and investigate guarantees on the misclassification error of the loss-minimizer in terms of the margin error rate of the best predictor. We show that, aiming for such a guarantee, the hinge loss is essentially optimal among all convex losses.

## 1. Introduction

Perhaps the most fundamental question studied in the theory of machine learning is that of binary classification with half-spaces. However the problem of agnostically learning half-spaces is known to be NP-hard in general (Kearns et al., 1994). Even when one only wants to learn a half-space relative to the best possible $M$-margin error, Ben-david & Simon (2000) show that (subject to $P \neq NP$) there exists no proper learning algorithm (i.e. returning a linear predictor) that runs in time polynomial in both $1/M$ and the desired accuracy. Under a cryptographic hardness assumption, (Shalev-Shwartz et al., 2010) extend this result to improper learning (i.e. when the algorithm may output any predictor, as long as it generalizes well).

In practice, one typically reverts to minimizing a convex surrogate, such as the hinge-loss, squared loss, logistic loss, exp-loss etc. Minimizing such a convex loss is usually easy, and can be done in polynomial time, but the question then is how well does minimizing such a convex surrogate perform relative to minimizing the actual classification error. An important line of work focused on relating the *excess* loss to the *excess* misclassification error, and introducing the notion of "classification calibrated" loss functions (Zhang, 2004; Bartlett et al., 2006). As discussed in detail in Section 4, this notion is relevant only when the Bayes predictor is in our hypothesis class—i.e. for linear prediction, when the Bayes predictor is exactly linear. Here we consider the more realistic agnostic setting, where we make no such assumption. Instead, we only rely on the existence of some linear predictor with small error rate at some margin, and ask the question of what misclassification error can be guaranteed by minimizing a convex loss. We obtain guarantees for specific loss functions, which allow us to compare between them, as well as a lower bound that holds for *any* convex loss. We now proceed to discuss and compare our work with other related work.

### 1.1. More Related Work

Several other authors, beyond those discussed in the previous section, also address the question of choosing appropriate surrogate loss functions for binary classification. Masnadi-Shirazi & Vasconcelos (2009) and Nock & Nielsen (2008) study various choices of surrogate losses for the classification problem and argue that the "right" loss depends on the underlying un-known distribution. In the agnostic setting, which we consider here, we do not know the underlying distribution, and so seek a distribution free guarantee. Christmann & Steinwart (2004) investigate robustness properties of learning algorithms that are based on convex risk minimization in terms of the so called influence function corresponding to the loss. They argue that most statistical models are intended to be approximations of true model generating data. Robustness of the method basically implies that when the true underlying distributions deviate from the model only slightly then the performance of the method is not affected much. Hence, they use this to argue about the efficacy of algorithms based on convex losses. While robustness property tells us that deviating from the model by a small amount doesn't affect performance much, it does not address the issue of agnostic learning where no assumptions are made about the underlying distribution (except that a hypothesis in the hypothesis class has low risk). In contrast, our work directly looks at how minimizing a surrogate loss corresponds to minimizing the misclassification error rate without any assumptions about the model generating the data or assuming even that the approximate Bayes optimal, w.r.t. the surrogate loss, belongs to the hypothesis class.

In terms of directly minimizing the misclassification error rate, without using a convex surrogate loss, Kalai et al. (2008) provide non-asymptotic finite sample bounds for efficient binary prediction with half spaces. However, to do so they assume the inputs are uniformly distributed on the surface of a sphere, which is an extremely strong an unrealistic assumption. Similarly, Kalai & Sastry (2009) also provide an efficient algorithm for minimizing the misclassification error, but only under the assumption that the conditional distribution of the label given the input $x$ is some monotonic function of $w \cdot x$ for some $w$—again significantly departing from the agnostic setting.

As mentioned earlier, although they focus on boosting (coordinate descent optimization), Long & Servedio (2008) essentially establish that if one does not assume that margin error, $\nu$, of the optimal linear classifier is small enough then any algorithm minimizing any convex loss $\phi$ (which they think of as a "potential") can be forced to suffer a large misclassification error. They do not, however, consider the bounded-margin case, that is when the margin error $\nu$ of the best linear classifier is small compared to the margin $M$. In concurrent work, (Long & Servedio, 2011) do show that using convex surrogate losses for learning can at best only guarantee that the zero-one is bounded by $O(\nu/M)$, which is a result very similar to our Theorem 4, though this a much higher constant. They do not however show lower bounds for specific loss functions which gives one the tool to compare various convex surrogate losses in a precise manor. More interestingly, they provide a randomized *improper* learning algorithm whose zero-one loss is bounded by $\nu/(M \log(\frac{1}{M})) + \epsilon$ in time polynomial in $1/M$ and $1/\epsilon$ (where $\nu$ is the misclassification error rate at margin $M$). This shows that, at least when improper learning is allowed (i.e. we are allowed to return a non-linear predictor, as long as it generalizes well), it is possible to do (slightly) better than minimizing a convex surrogate. The question of whether this is possible with proper learning remains open.

## 2. Setting

Let $\mathcal{D}(\mathbf{x}, y)$ be a distribution over $U \times \{+1, -1\}$, where, for some $d$, $U = \{\mathbf{x} \in \mathbb{R}^d : \|\mathbf{x}\| \leq 1\}$ is the $d$-dimensional unit sphere. We will often actually consider finite samples, in which case $\mathcal{D}$ should be understood as a uniform distribution over points in the sample.

A linear predictor is described by a vector and a bias term: $(\mathbf{w}, w_0)$, $\mathbf{w} \in \mathbb{R}^d$, $w_0 \in \mathbb{R}$. For a loss function $\phi : \mathbb{R} \to \mathbb{R}$, the $\phi$-risk is give by:

$$R_\phi^\mathcal{D}(\mathbf{w}, w_0) = \mathbb{E}_{(\mathbf{x},y)\sim\mathcal{D}}\left[\phi(y(\langle\mathbf{w}, \mathbf{x}\rangle + w_0))\right].$$

When the distribution $\mathcal{D}$ is understood from the context, we will simply use $R_\phi(\mathbf{w}, w_0)$. We will be particularly interested in the 0-1 loss $\phi_{01}(z) = \mathbf{1}\{z \leq 0\}$ and the margin-loss $\phi_m(z) = \mathbf{1}\{z < 1\}$, and the corresponding risks:

$$R_{01}(\mathbf{w}, w_0) = R_{\phi_{01}}(\mathbf{w}, w_0) = \Pr\left(y(\langle\mathbf{w}, \mathbf{x}\rangle + w_0) \leq 0\right)$$

$$R_m(\mathbf{w}, w_0) = R_{\phi_m}(\mathbf{w}, w_0) = \Pr\left(y(\langle\mathbf{w}, \mathbf{x}\rangle + w_0) < 1\right)$$

Note that we are considering a prediction of zero as an error both for the positive class and the negative class, thus always predicting zero yields an error rate of one. We are also using $\phi_m$ and $R_m$ to denote the error relative to a margin of 1, and so we will actually encode the margin through the norm of $\mathbf{w}$, i.e. the actual margin is $1/\|\mathbf{w}\|$.

## 3. Misclassification Error Guarantee

We begin by showing that for any convex loss function, and any arbitrarily low misclassification error rate $\nu > 0$, there exists a sample which is linearly separable with error rate $\nu$, but for which the predictor minimizing the surrogate loss would have error rate arbitrarily close to 1. In other words, there are training samples over which an algorithm that minimizes the surrogate loss will output a classifier whose actual training error is close to 1, in spite of the fact that these samples can be classified with small error by another half-space (which is missed by the loss minimization algorithm, due to having high surrogate loss). Furthermore, such examples exist even if the domain space is just the real interval. A similar result was also essentially shown by Long & Servedio (2008)— they discuss "boosting", i.e. loss minimization by coordinate descent, here we refer more directly to the loss minimizer.

For a loss $\phi(\cdot)$, define the *Misclassification* **E***rror* **G***uarantee* relative to the zero-one loss as:

$$\mathbf{EG}(\phi, \nu) = \sup_{\substack{\mathcal{D} \text{ s.t. } \exists \mathbf{w}, w_0, \\ R_{01}(\mathbf{w}, w_0) \leq \nu}} \sup_{\substack{(\hat{\mathbf{w}}, \hat{w}_0) \text{ s.t. } \forall \mathbf{w}, w_0 \\ R_\phi(\hat{\mathbf{w}}, \hat{w}_0) \leq R_\phi(\mathbf{w}, w_0)}} R_{01}(\hat{\mathbf{w}}, \hat{w}_0) \quad (1)$$

That is, we are asking: what is the largest misclassification error suffered by the linear predictor minimizing $\phi$-risk when the underlying distribution is such that the misclassification error rate of the best half-space is bounded by $\nu$.

When $\nu = 0$, i.e. in the separable case, this is essentially a question about "classification calibration", and we have that for convex $\phi$, $\mathbf{EG}(\phi, 0) = 0$ if and only if $\phi$ is differentiable at zero and $\phi'(0) < 0$ (Bartlett et al., 2006)—see Section 4. This is a mild condition that holds for most common loss functions, but here we are interested in $\mathbf{EG}(\phi, \nu)$ for $\nu > 0$.

Our initial claim can be state as follows:

**Proposition 1.** *For any convex function $\phi$, and any $\nu > 0$, $\mathbf{EG}(\phi, \nu) = 1$.*

This follows from the one-dimensional source distribution below:

- There are $\nu/2$ points at $x = -1$ labelled $+1$
- There are $\nu/2$ points at $x = +1$ labelled $-1$
- There are $\frac{1-\nu}{2}$ points at $x = M$ labelled $+1$
- There are $\frac{1-\nu}{2}$ points at $x = -M$ labelled $-1$

Here, the optimal linear predictor is one that labels points to the right of 0 as positive and left of 0 as negative, yielding misclassification error $\nu$. However as $M \to 0$, the minimizer of any classification calibrated convex surrogate loss will label points to the right of 0 as negative and left of 0 as positive because the points close to 0 ($M$ close) suffer small loss under the convex loss. This simple example shows that $\mathbf{EG}(\phi, \nu) = 1$ for any convex loss.

### 3.1. Surrogate loss minimization when margins exist

We have just shown that the existence of a linear predictor with low misclassification error is not enough to ensure the success of surrogate loss minimization. Our next step is to analyze the success of this paradigm under stronger assumptions - the existence of a good linear classifier with respect to some positive margin. Our next definition considers the worst possible error of a surrogate loss minimizer when the data allows a low error classifier with some margin. Our main result in this section, Theorem 4, implies that, up to a factor of 2, the hinge loss is optimal among all convex loss functions in that respect.

$$\mathbf{EG}(\phi, \nu, B)$$
$$= \sup_{\substack{\mathcal{D} \text{ s.t.} \\ \exists \mathbf{w}, w_0, \|\mathbf{w}\| \leq B, \\ R_m(\mathbf{w}, w_0) \leq \nu}} \sup_{\substack{(\hat{\mathbf{w}}, \hat{w}_0) \text{ s.t. } \forall \mathbf{w}, w_0, \\ R_\phi(\hat{\mathbf{w}}, \hat{w}_0) \leq R_\phi(\mathbf{w}, w_0)}} R_{01}(\hat{\mathbf{w}}, \hat{w}_0) \quad (2)$$

Here, $B$ specifies the margin, and we will sometimes refer to it directly as $M = 1/B$. We have that $\mathbf{EG}(\phi, \nu) = \mathbf{EG}(\phi, \nu, \infty)$. A more careful look at the lower bound on $\mathbf{EG}(\phi, \nu)$ in the previous section reveals that for $\nu \geq 1/(B+1) = M/(M+1)$, we have $\mathbf{EG}(\phi, \nu, B) = 1$ for any convex loss $\phi$. However for smaller values of $\nu$, it is possible to get meaningful bounds on $\mathbf{EG}(\phi, \nu, B)$. In particular, for the hinge loss the simple observation that $B + 1$ times the margin loss upper bounds the hinge loss in the interval $[-B, B]$ gives the below upper bound on $\mathbf{EG}(\phi, \nu, B)$.

**Proposition 2.** *For the hinge loss $\phi_{hinge}(z) = \max(0, 1 - z)$, we have that*

$$\mathbf{EG}(\phi_{hinge}, \nu, B) \leq (B+1)\nu$$

In order to prove our main result, it would be useful to generalize the above result a parametric family of "scaled" hinge losses give by $\phi_{\gamma-\text{hinge}}(z) = \max(0, 1 - \frac{z}{\gamma})$, with a parameter $\gamma > 0$. Thus, if $\gamma = 1$, $\phi_{\gamma-\text{hinge}}$ is $\phi_{\text{hinge}}$.

**Theorem 3.** *For all $\gamma > 0$, if $\nu(B + 1) < 1$, then*

$$\mathbf{EG}(\phi_{\gamma-hinge}, \nu, B) \geq \min\left\{\frac{\nu(B+1)}{2}, 1 - 2\nu\right\}$$

Using the above theorem, we prove our main result:

**Theorem 4.** *For any convex loss function $\phi$, we have*

$$\mathbf{EG}(\phi, \nu, B) \geq \min\left\{\frac{\nu(B+1)}{2}, \frac{1}{2}\right\}$$

The above theorem and Proposition 2 together show that hinge loss is optimal up to constant factor 2.

We would also like to compare the error guarantees of specific losses or loss families. To this end, we derive the following generic "recipe" for obtaining lower bounds specific to loss functions (See Appendix, Section A for a proof):

**Lemma 3.1.** *For any non-negative convex loss $\phi$,*

$$\boldsymbol{EG}(\phi, \nu, B) \geq \qquad (3)$$
$$\geq \min \left\{ \sup_{\beta \in [0,1]} \inf_{\alpha \in [0, \frac{4}{\sqrt{5}}]} \frac{(1-\nu)\phi(\alpha) + \nu \phi\left(-(B-5)\frac{\alpha}{2}\right) - \phi(2\beta)}{2(\phi(-2\beta) - \phi(2\beta))}, \frac{1}{4} \right\}$$

Based on the above lemma, we show a lower bound for any strongly convex loss function, which shows that choosing strongly convex loss functions is in fact qualitatively worse in the worst case sense (See Appendix, Section B for a proof).

**Corollary 5.** *For any $\lambda$-strongly convex surrogate loss that is L-Lipschitz in the interval $[-1, 1]$, we have that*

$$\boldsymbol{EG}(\phi, \nu, B) \geq \min\left\{ \frac{\lambda}{64L} \nu (B-1)^2, \frac{1}{16} \right\}$$

### 3.2. Bounds for Specific Losses

Before we proceed we would like to give an alternate bound to Equation (3) which is often easier to get a handle on. To this end note that by (3.1) :

$$\boldsymbol{EG}(\phi, \nu, B) \geq \min \left\{ \frac{1}{4}, \right.$$
$$\left. \sup_{\beta \in [0,1]} \inf_{\alpha \in [0, \frac{4}{\sqrt{5}}]} \max \left\{ \frac{\phi(\alpha) - \phi(2\beta)}{4\phi(-2\beta) - \phi(2\beta)}, \frac{\nu\left(\phi\left(-(B-5)\frac{\alpha}{2}\right) - \phi(2\beta)\right)}{2(\phi(-2\beta) - \phi(2\beta))} \right\} \right\}$$

For the first term in the max, note that for some fixed $x_2$ the ratio $\frac{\phi(x_1) - \phi(x_2)}{x_1 - x_2}$ is monotonically non-decreasing in $x_1$ and so,

$$\boldsymbol{EG}(\phi, \nu, B)$$
$$\geq \min \left\{ \frac{1}{4}, \sup_{\beta \in [0,1]} \inf_{\alpha \in [0, \frac{4}{\sqrt{5}}]} \max \left\{ \frac{2\beta - \alpha}{8\beta}, \frac{\nu\left(\phi\left(-(B-5)\frac{\alpha}{2}\right) - \phi(2\beta)\right)}{2(\phi(-2\beta) - \phi(2\beta))} \right\} \right\}$$

Further note that for $\beta > \alpha$, $\frac{2\beta - \alpha}{4\beta} \geq \frac{1}{4}$ and so we can conclude that,

$$\boldsymbol{EG}(\phi, \nu, B) \geq \min \left\{ \sup_{\beta \in [0,1]} \frac{\nu\left(\phi\left(-(B-5)\frac{\beta}{2}\right) - \phi(2\beta)\right)}{2(\phi(-2\beta) - \phi(2\beta))}, \frac{1}{8} \right\} \quad (4)$$

**Example 3.1** (Hinge Loss). *The hinge loss is give by $\phi(z) = \max(1 - z, 0)$. Simply using Theorem 4 we get*

$$\boldsymbol{EG}(\phi, \nu, B) \geq \min \left\{ \frac{\nu(B+1)}{2}, \frac{1}{2} \right\}$$

**Example 3.2** (Squared Hinge Loss). *The squared hinge loss is given by $\phi(z) = \max(1 - z, 0)^2$. Using Equation 4 with $\beta = \frac{1}{2}$, we get*

$$\boldsymbol{EG}(\phi, \nu, B) \geq \min \left\{ \frac{\nu(B-1)^2}{128}, \frac{1}{8} \right\}$$

**Example 3.3** (Exponential Loss). *Exponential loss is given by $\phi(z) = e^{-z}$. For Exponential loss using Equation 4 with $\beta = 1/2$ we get*

$$\boldsymbol{EG}(\phi, \nu, B) \geq \min \left\{ \frac{\nu(e^B - 1)}{2(e^2 - 1)}, \frac{1}{8} \right\}$$

**Example 3.4** (Logistic Loss). *For Logistic loss is given by $\phi(z) = \log(1 + e^{-z})$. For logistic loss, using Equation 4 with $\beta = 1/2$ we get,*

$$\boldsymbol{EG}(\phi, \nu, B) \geq \min \left\{ \frac{\nu \left( \log(1 + e^{(B-5)/4}) - 0.32 \right)}{2}, \frac{1}{8} \right\}$$

Notice that for large $B$ this behaves similar to hinge loss. Also notice that the squared hinge loss (and similarly square loss) behave quadratically in $B$ and exponential loss has exponential dependence on $B$. Thus we see that for large $B$ hinge loss gives qualitatively better bound that squared loss of exponential loss.

### 3.3. General Hypothesis Classes

The misclassification error guarantee (**EG**) was specific to linear predictors (with norm bounded by $B$). One can easily generalize this definition of misclassification error guarantee w.r.t. an arbitrary hypothesis class $\mathcal{H}$ as follows :

$$\boldsymbol{EG}^{\mathcal{H}}(\phi, \nu, B)$$
$$= \sup_{\substack{\mathcal{D} \text{ s.t. } \exists h \in \mathcal{H}, \\ \sup_{\mathbf{x}, \mathbf{x}'} |h(\mathbf{x}) - h(\mathbf{x}')| \leq 2B, \\ R_m(h) \leq \nu}} \sup_{\substack{\widehat{h} \in \mathcal{H} \text{ s.t.} \\ \forall h \in \mathcal{H}, R_\phi(\widehat{h}) \leq R_\phi(h)}} R_{01}(\widehat{h}) \quad (5)$$

Since the linear hypothesis with norm bounded by $B$ is a particular case of a hypothesis class that satisfies $\sup_{\mathbf{x}, \mathbf{x}'} |\langle \mathbf{w}, \mathbf{x} \rangle - \langle \mathbf{w}, \mathbf{x}' \rangle| \leq 2B$, we have that for any loss $\phi$ :

$$\boldsymbol{EG}(\phi, \nu, B) \leq \sup_{\mathcal{H}} \boldsymbol{EG}^{\mathcal{H}}(\phi, \nu, B)$$

On the other hand, note that Proposition 2 was only based on the fact that hinge loss can be bounded by the margin loss times the maximum value of the loss (given maximal value of predictor is $B$). Hence the proposition directly extends to any hypothesis class and so we can conclude that

$$\sup_{\mathcal{H}} \boldsymbol{EG}^{\mathcal{H}}(\phi_{\text{hinge}}, \nu, B) \leq \nu(B + 1). \quad (6)$$

And so, all our upper and lower bounds can also be interpreted as upper and lower bounds on $\sup_{\mathcal{H}} \mathbf{EG}^{\mathcal{H}}(\phi_{\text{hinge}}, \nu, B)$, i.e. on the best misclassification error guarantee that is possible based only on the loss function, and is required to hold independent of the hypothesis class.

Furthermore, by Theorem 4, the right hand side in (6) is in turn bounded by $2\mathbf{EG}(\phi_{\text{hinge}}, \nu, B)$, and we see that the "extreme" hypothesis class for the hinge-loss is the linear class. This can also be extended to the other commonly used losses referred to in Section 3.2. Hence, even if we want misclassification error guarantees that focus only on the loss and are required to hold regardless of the hypothesis class, studying the linear class, i.e. $\mathbf{EG}(\phi, \nu, B)$, is often sufficient.

The main reason we focus our presentation on linear predictors is that learning linear predictors (possibly linear in a feature space, which includes kernel methods) combined with convex losses is essentially the only situation that yields a convex optimization problem, which is one of our goals when using convex surrogates.

## 4. Classification Calibration and the Misclassification Error Guarantee

An important line of work focused on relating the *excess* loss to the *excess* misclassification error, and introducing the notion of "classification calibrated" loss functions (Zhang, 2004; Bartlett et al., 2006). A basic notion here is that of a loss being "classification calibrated", i.e. ensuring that zero excess loss (beyond the Bayes optimal) translates to zero excess misclassification error (beyond the Bayes optimal). This ensures that *if we consider a class rich enough to include the Bayes optimal predictor* then minimizing the expected loss indeed also minimizes the misclassification error. Classification calibration can be seen as an extreme point of the misclassification error guarantee ($\mathbf{EG}$) in two ways:

First, for any convex loss $\phi$, $\mathbf{EG}(\phi, 0) = 0$ if and only if $\phi$ is classification calibrated (both are equivalent to the derivative at zero being defined and negative). Discussing $\nu = 0$ corresponds to considering only the separable case, which is in a sense the point of intersection of our study and that of Zhang (2004); Bartlett et al. (2006).

Second, we can think of classification calibration as referring to $\mathbf{EG}^{\mathcal{M}}$, where $\mathcal{M}$ is the set of all measurable functions. That is, a surrogate loss $\phi$ is classification calibrated if and only if $\mathbf{EG}^{\mathcal{M}}(\phi, nu) = \nu$.

Analyzing either $\mathbf{EG}(\phi, 0)$ or $\mathbf{EG}^{\mathcal{M}}(\phi, \nu)$ is not satisfactory as they don't correspond to the agnostic learning case where we are interested in doing as well as the best hypothesis in the function class of interest. Typically, classification calibration based results are used in conjunction with approximation theory to argue that as the number of training samples increase one can consider richer and richer hypothesis classes, and hence eventually converge to the set of all measurable functions $\mathcal{M}$, where $\mathbf{EG}^{\mathcal{M}}(\phi, \nu)$, and hence the notion of classification calibration, is relevant. However, such an analysis typically only establishes asymptotic behavior (unless strong assumptions are made). The analysis in this work neither needs to assume that data is linearly separable nor assume that the Bayes optimal predictor under the surrogate loss function is linear (with norm bounded by $B$).

For example, based on Zhang (2004), Rosasco et al. (2004) argue that the hinge loss (and also logistic loss) enjoy better rates than other losses like squared loss. However these results are also based on convergence to the Bayes optimal and so we either need to take very rich hypothesis classes or assume that the Bayes optimal predictor under surrogate loss is contained in the hypothesis class used.

## 5. Including Estimation Error Rates

It is interesting to consider how misclassification error guarantee ($EG$) combines with estimation error rate. In practice, we get a finite training sample and when picking the hypothesis that minimizes some empirical loss, the estimation error involved in minimizing empirical objective, rather than the true expected objective, comes into the picture. While choosing the loss $\phi$ one should take into account both the misclassification error guarantee ($\mathbf{EG}$) associated with the loss and also the associated estimation error for the problem. For example, thinking of only estimation error, one might think that squared error is better as one might expect a $1/n$ rate where $n$ is the sample size. However, $\mathbf{EG}$ for squared loss is large as we argued in Section 3.2.

For high dimensional cases, one can argue that hinge loss is the loss of choice even when we take estimation error into account. For the conservative update algorithm w.r.t. the hinge loss with its corresponding analysis by (Shalev-Shwartz, 2007), or for the exact minimizer (which corresponds to the SVM) of empirical hinge loss using results in (Srebro et al., 2010) (and noticing that hinge loss upper bounds a smooth version of margin loss which in turn upper bounds the zero-one loss) one can show that if $\widehat{\mathbf{w}}_n$ is the linear predictor returned by one of these algorithms, then, in expectation over training sample

$$R_{01}(\hat{\mathbf{w}}_n) \leq 2 \inf_{\mathbf{w}:\|\mathbf{w}\|\leq B} R_{\phi_{\text{hinge}}}(\mathbf{w}) + O\left(\frac{B^2}{n}\right)$$

$$\leq 2\,(B+1)\nu + O\left(\frac{B^2}{n}\right) \qquad (7)$$

where $\nu = \inf_{\mathbf{w}:\|\mathbf{w}\|\leq B} R_{01}(\mathbf{w})$. In order to compare this with the squared loss, we first note that the best misclassification error guarantee one can give for an algorithm that minimizes the expected squared loss is bounded by $\mathbf{EG}(\phi_{\text{squared}}, \nu, B) \geq \min\{\nu(B-1)^2/128, 1/8\}$. Further, when the dimensionality is large (compared to the sample size) then the estimation error rate for the squared loss (with linear predictors) can be lower bounded by $B^2/n$ (see for instance (Srebro et al., 2010)) and so the best guarantee that can be provided on the classification risk of estimator obtained by minimizing squared loss scales is $\nu(B-1)^2 + B^2/n$. Comparing this with the upper bound in equation 7 shows that the hinge loss is qualitatively superior (in the worst case) even if one takes into account the estimation error rates.

A similar analysis can be repeated w.r.t. other losses where effectively hinge loss (and also logistic loss) can be shown to have qualitatively better performance than, for instance, squared loss or exponential loss, or any strongly convex loss. We would like to point out that the low-dimensional analysis requires a bit more care, as the estimation error for the squared loss might be much lower than for methods based on other loss functions.

## 6. Proofs of Theorems 3 and 4

*Proof of Theorem 3.* The distribution for this theorem is as follows:

- There are $\nu$ points at $x = -1$ labelled $+1$
- There are $\beta$ points at $x = -M$ labelled $-1$
- There are $1 - \beta - \nu$ points at $x = M$ labelled $+1$

Although the data lies on the real line, we consider the example to be in $\mathbb{R}^2$. Now consider the classifier found when minimizing the convex surrogate loss $\phi_{M-\text{hinge}}$. We aim to show that the vector $\mathbf{w}^\star = (0, -1)$ with $\mathbf{w}_0^\star = M$ is the optimal classifier. It has $R_\phi(\mathbf{w}^\star, \mathbf{w}_0^\star) = 2\beta$. The margin loss of $(\mathbf{w}^\star, \mathbf{w}_0^\star)$ is clearly $\beta$. Any other classifier that misclassifies fewer than $\beta$ points, must cross the $x$-axis.

### Case 1

Assume that we have a classifier that misclassifies fewer than $\beta$ points, and that this classifier intersects the $x$-axis between $[-M, 0]$ and mislabels only (and all) $\nu$ points at $x = -1$. Assume that it crosses at $-c \in [-M, 0]$. Assume also that it crosses the $x$-axis with some angle $\theta$. Thus, $\mathbf{w} = (sin(\theta), -cos(\theta))$ and $w_0 = sin(\theta)c$.

Consider the case where $(\mathbf{w}, w_0)$ is further than $M$ away from the $1 - \beta - \nu$ points at $x = M$. Then, $R_\phi(\mathbf{w}, w_0) = \nu(1 + \frac{sin(\theta)(1-c)}{M} + \beta(1 - \frac{sin(theta)(M-c)}{M})$. By taking the derivative, we can see this is increasing as $\theta$ increases, thus taking $\theta$ to be as small as possible while maintaining the $M$ distance from the points at $x = M$ is best. This gives, $sin(\theta) = \frac{M}{M+c}$, which yields $R_\phi(\mathbf{w}, w_0) = \frac{1}{M+c} \cdot (\nu(1 + M) + 2c\beta)$.

Finally, taking the derivative with respect to $c$ we find that if $\beta < \nu \frac{1+M}{2M}$ then $R_\phi(\mathbf{w}, w_0)$ is minimized at $c = M$ with a cost of $R_\phi(\mathbf{w}, w_0) = \nu \frac{1+M}{2M} + \beta$.

We do not need to consider similar classifiers that intersect the $x$-axis between $(0, M]$, as they will only increase the cost of points at $x = -1$ and add cost of points from $x = M$. This will never beat the classifier mentioned above when $c = 0$, which is already beaten by the classifier with $c = M$.

### Case 2

Now assume that we have a classifier that misclassifies fewer than $\beta$ points and that this classifier intersects the $x$-axis between $[-\frac{1+M}{2}, -M]$, and that the classifier mislabels only (and all) $1 - \beta - \nu$ points at $x = M$. Assume the classifier crosses the $x$-axis at position $-c$, and that it crosses with some angle $\theta$. Thus, $\mathbf{w} = (-sin(\theta), cos(\theta))$ and $w_0 = sin(\theta)c$.

If the classifier is at least distance $M$ from all of the points, then $R_\phi(\mathbf{w}, w_0) \geq (1 - \beta - \nu)\frac{2c}{c-M}$, which is minimized when $c$ is largest. Thus, $R_\phi(\mathbf{w}, w_0) \geq 2(1 - \beta - \nu)\frac{1+M}{1-M}$. Note that $2\beta < 2(1 - \beta - \nu)\frac{1+M}{1-M}$ if $\nu < \frac{M}{1+M}$, which is one of our assumptions. So, for this case, $(\mathbf{w}, w_0)$ is not optimal.

Finally, for this case, we have where the points at $x = -M$ are within the margin of the classifier, and so they contribute to the $\phi$-loss. Let $c' = c - M$. This gives $R_\phi(\mathbf{w}, w_0) = \beta(1 - \frac{sin(\theta)c'}{M}) + (1 - \beta - \nu)(1 + \frac{sin(\theta)(c'+2M)}{M})$.

Using the derivative, we find that it is minimal when $sin(\theta) = \frac{M}{1-c'-M}$. This yields $R_\phi(\mathbf{w}, w_0) = \frac{1}{1-c'-M}(-2\beta(c' + M) + (1 - \nu)(1 + M))$. Taking the derivative with respect to $c'$, yields the minimum when $c' = 0$ (which means $c = M$). Therefore, the cost is $R_\phi(\mathbf{w}, w_0) = \frac{1}{1-M}(-2\beta M + (1 - \nu)(1 + M))$. Again, we find that this cost is larger than $2\beta$ when $\nu < \frac{M}{1+M}$.

Provided $1 - \beta - \nu > \nu$, we do not need to consider similar classifiers that intersect the $x$-axis between $[-1, -\frac{1+M}{2})$, as they always have higher cost.

**Remaining cases**

Any classifiers outside of those discussed in Case 1 and Case 2 misclassify at least $\beta$ points, as long as $\beta < \frac{1}{2}$.

Therefore, since $R_\phi(\mathbf{w}^\star, \mathbf{w}_0^\star) = 2\beta$ we merely need that $2\beta < \nu\frac{1+M}{2M} + \beta$ to handle Case 1, which is true when $\beta < \nu\frac{1+M}{2M}$. This was our assumption.

Finally, the $\mathbf{EG}(\cdot)$ found here applies to all $\phi_{\gamma-\text{hinge}}$ losses because $\phi_{M-\text{hinge}}(x) = \phi_{\gamma-\text{hinge}}(\frac{\gamma}{M}x)$, and thus they are equivalent losses with respect to their error guarantee. Thus what we have shown is that $\mathbf{EG}(\phi_{\gamma-\text{hinge}}, \nu, B) \geq \beta$ for any $\beta$ such that $\nu < \beta < \frac{\nu(B+1)}{2}$ and $1 - \beta > 2\nu$. Thus taking the largest of such $\beta$'s gives the final form of the bound. $\square$

*Proof of Theorem 4.* There exists an $\alpha$, with $0 < \alpha$ such that the horizontal line above the x-axis, labelling all points as $+1$, has cost $\beta \cdot \phi(-\alpha) + (1-\beta) \cdot \phi(\alpha) < 1$, as long as $\beta < \frac{1}{2}$, which holds for this theorem since $\beta < \nu\frac{M+1}{2M}$.

For any convex function $\phi$, there exists a $\gamma > 0$ such that $\phi(x) \geq \phi_{\gamma-\text{hinge}}(x)$, for all $x$. Thus, any classifier $(w, w_0)$ that misclassifies fewer than $\beta$ points must satisfy $R_\phi(w, w_0) \geq \beta + \frac{\nu}{2} \cdot (1+B)$. This follows from the proof of Theorem 3, because it is the minimum value of $R_{\phi_{\gamma-\text{hinge}}}(w, w_0)$, where $(w, w_0)$ misclassifies fewer than $\beta$ points.

However, for any classifiers $(w, w_0)$ that misclassifies fewer than $\beta$ points, it is some distance $c$ away from at least $\nu$ points that it misclassifies, where $c > 0$. This follows from our distribution having three groups of points: a group of size $\nu$, a group of size $\beta$, and a group of size $1 - \beta - \nu$. The smallest group size is $\nu$ because $\beta > \nu$ and $1 - \beta > 2\nu$ and that $(w, w_0)$ must misclassify at least one of the groups of points. Therefore, $R_\phi(w, w_0) \geq \beta + \frac{\nu}{2} \cdot (1+B) + \nu(\phi(-c) - \phi_{\gamma-\text{hinge}}(-c))$.

Note that $\phi(-c) - \phi_{\gamma-\text{hinge}}(-c) \geq 0$. From here, we break the analysis into two cases. First, for all $x \leq 0$, $\phi(x) = \phi_{\gamma-\text{hinge}}(x)$. Second, is where there exists $x' < 0$ such that $\phi(x') > \phi_{\gamma-\text{hinge}}(x')$.

For both cases, we will make use of the cost of a horizontal line above the x-axis that labels all points as $+1$. Recall from above that the cost of this classifier, which is $(0, w_0')$ for some $w_0' > 0$, is given by $\beta \cdot \phi(-\alpha) + (1-\beta) \cdot \phi(\alpha) < 1$. Also, this inequality holds for any $\phi(\cdot)$. We refer to this classifier as $h^*$ and its risk is $R_\phi(h^*)$.

**Case 1:**

In this case, $\phi(-c) - \phi_{\gamma-\text{hinge}}(-c) = 0$ for all $c > 0$. Thus, we are back to $R_\phi(w, w_0) \geq \beta + \frac{\nu}{2} \cdot (1+B)$. However, we can simplify $R_\phi(h^*)$ in this case.

$$R_\phi(h^*) = \beta \cdot \phi(-\alpha) + (1-\beta) \cdot \phi(\alpha)$$
$$= \beta \cdot \phi_{\gamma-\text{hinge}}(-\alpha) + (1-\beta) \cdot \phi(\alpha)$$

Now, since $R_\phi(w, w_0) \geq \beta + \frac{\nu}{2} \cdot (1+B)$ for any convex loss function $\phi(\cdot)$, we can replace $\phi(x)$ with $\phi'(x) = \phi(kx)$ for any $k > 0$ and the same inequality must hold for $\phi'(x)$. Further, $\phi'(x) = \phi_{\gamma-\text{hinge}}(kx)$, for all $x \leq 0$ and $k > 0$. Recall that as $x \to \infty$, $\phi'(x) \to 0$. Let $\phi'_{\gamma-\text{hinge}}(x) = \phi_{\gamma-\text{hinge}}(kx)$.

$$R_{\phi'}(h^*) = \beta\phi'(-\alpha) + (1-\beta)\phi'(\alpha)$$
$$= \beta\phi_{\gamma-\text{hinge}}(-k\alpha) + (1-\beta)\phi(k\alpha)$$
$$= \beta\phi'_{\gamma-\text{hinge}}(-\alpha) + (1-\beta)(\phi'_{\gamma-\text{hinge}}(\alpha) + \epsilon_{\phi'})$$
$$= R_{\phi'_{\gamma-\text{hinge}}}(h^*) + (1-\beta)\epsilon_{\phi'}$$
$$< R_{\phi'_{\gamma-\text{hinge}}}(h^*) + \epsilon_{\phi'}$$

As $k \to \infty$, $\epsilon_{\phi'} \to 0$. Finally, we have from Theorem 3, that, for all $k > 0$, $R_{\phi'_{\gamma-\text{hinge}}}(h^*) < \beta + \frac{\nu}{2} \cdot (1+B)$. This implies that there exists an $\epsilon > 0$ such that $R_{\phi'_{\gamma-\text{hinge}}}(h^*) + \epsilon < \beta + \frac{\nu}{2} \cdot (1+B)$. Take $k$ to be large enough such that $\epsilon_{\phi'} \leq \epsilon$. This implies that $R_{\phi'}(h^*) < R_{\phi'_{\gamma-\text{hinge}}}(h^*) + \epsilon < \beta + \frac{\nu}{2} \cdot (1+B)$.

Therefore, there exists a $k > 0$ such that for $\phi'(x) = \phi(kx)$, $\mathbf{EG}(\phi'(x), \nu, B) \geq \beta$.

Finally, because $\phi'(x) = \phi(kx)$, where $k > 0$, they are equivalent losses. Therefore, $\mathbf{EG}(\phi(x), \nu, B) \geq \beta$.

**Case 2:**

Consider loss $\phi'(x) = \phi(kx)$, where $k > 0$ is large enough such that $\phi(-kc) - \phi_{\gamma-\text{hinge}}(-kc) > 1 - \beta - \frac{\nu}{2} \cdot (1+B)$. This is possible since $\phi(\cdot)$ is convex and we assumed that for some $x < 0$, $\phi(x) > \phi_{\gamma-\text{hinge}}(x)$.

Therefore, $R_{\phi'}(w, w_0) > 1$. But we know that there exists $\alpha > 0$ such that $(0, w_0')$, with $w_0' > 0$ (i.e. a horizontal line above the x-axis that labels all points as $+1$) has cost $\beta \cdot \phi'(-\alpha) + (1-\beta) \cdot \phi'(\alpha) < 1$. Thus, for $\phi'(\cdot)$, $\mathbf{EG}(\phi'(\cdot), \nu, B) \geq \beta$.

Finally, because $\phi'(x) = \phi(kx)$, where $k > 0$, they are equivalent losses. Therefore, if $0 < \nu < \frac{M}{M+1} = \frac{1}{B+1}$, then $\mathbf{EG}(\phi(\cdot), \nu, B) \geq \beta$, for all $\beta$ such that $\nu < \beta < \frac{\nu}{2} \cdot (1+B)$ and $1 - \beta > 2\nu$.

To conclude the proof note that as already mentioned, when $\nu(B+1) \geq 1$ we anyway get that $\mathbf{EG}(\phi, \nu, B) \geq 1$ and what we showed in the proof is that when $\nu(B+1) < 1$, then $\mathbf{EG}(\phi, \nu, B) \geq \beta$ for any $\beta$ s.t. $\beta < \frac{\nu(B+1)}{2}$ and $1 - \beta > 2\nu$. Since $B > 1$, $\nu < 1/2$ from which we conclude the proof. □

## 7. Discussion

In this paper, we provide lower bounds on the best misclassification error achievable by algorithms minimizing convex surrogate losses in terms of the $M$-margin error. Specifically, we show that the misclassification error rate of the linear predictor minimizing expected hinge loss is bounded by $\nu(B+1)$, where $\nu$ is the bound on the $M$-margin error and $B = 1/M$. Further, by showing that when using linear predictors any algorithm minimizing any convex loss has a misclassification error of at least $\frac{\nu(B+1)}{2}$, we conclude that the hinge loss is optimal up to factor 2. We also show lower bounds for specific convex losses and that any strongly convex loss has a qualitatively worse guarantee when compared to hinge loss. We argue that the analysis can be used to qualitatively compare convex surrogate losses used for binary classification, and show that the hinge loss is the loss of choice for classification problems. The relationship of the misclassification error guarantee term, which we introduce in this paper, with the notion of classification calibration of loss function is also explored. Specifically, we show how classification calibration can be seen as arising from an extreme case of our misclassification error guarantee term. As an example of the implications of our results, we argue that even when one takes estimation error rates into consideration, hinge loss is optimal up to constant factor (in the worst case sense, at least for high dimensional problems).